\documentclass[letterpaper, 10 pt, conference]{ieeeconf}
\usepackage{booktabs}
\usepackage{multicol, multirow}
\usepackage{placeins}
\usepackage{graphicx}
\usepackage{amsmath,amsfonts}
\usepackage{cite}
\usepackage{color}
\usepackage[colorlinks,
            linkcolor=red,
            anchorcolor=blue,
            citecolor=green
            ]{hyperref}
            
\usepackage{caption}
\usepackage{subcaption}

\usepackage{algorithm}
\usepackage{algorithmic}
\usepackage{threeparttable}
\newcounter{myfootnotecounter}

\IEEEoverridecommandlockouts
\DeclareMathOperator*{\argmin}{arg\,min}
\newcommand{\first}[1]{{\mathbf{#1}}}
\newcommand{\second}[1]{#1}
\newcommand{\third}[1]{#1}

\usepackage{amssymb}
\usepackage{pifont}

\title{\LARGE \bf
Domain Generalization for Vision-based Driving \\Trajectory Generation
}

\author{Yunkai Wang$^{1}$, Dongkun Zhang$^{1,2}$, Yuxiang Cui$^{1}$, Zexi Chen$^{1}$,\\ Wei Jing$^{2}$, Junbo Chen$^{2}$, Rong Xiong$^{1}$, Yue Wang$^{1}$$^{\dagger}$
\thanks{This work was supported by Alibaba Group through Alibaba Innovative Research (AIR) Program.}
\thanks{$^{1}$Yunkai Wang, Dongkun Zhang, Yuxiang Cui, Zexi Chen, Rong Xiong, and Yue Wang are with the State Key Laboratory of Industrial Control Technology and Institute of Cyber-Systems and Control, Zhejiang University, Hangzhou, China.}
\thanks{$^{2}$Dongkun Zhang, Wei Jing and Junbo Chen are with the Department of Autonomous Driving Lab, Alibaba DAMO Academy, Hangzhou, China}
\thanks{$^\dagger$ Corresponding author, {\tt\small wangyue@iipc.zju.edu.cn}}
}

\begin{document}

\maketitle
\thispagestyle{empty}
\pagestyle{empty}

\begin{abstract}
One of the challenges in vision-based driving trajectory generation is dealing with out-of-distribution scenarios.
In this paper, we propose a domain generalization method for vision-based driving trajectory generation for autonomous vehicles in urban environments, which can be seen as a solution to extend the Invariant Risk Minimization (IRM) method in complex problems.
We leverage an adversarial learning approach to train a trajectory generator as the decoder. Based on the pre-trained decoder, we infer the latent variables corresponding to the trajectories, and pre-train the encoder by regressing the inferred latent variable. Finally, we fix the decoder but fine-tune the encoder with the final trajectory loss.
We compare our proposed method with the state-of-the-art trajectory generation method and some recent domain generalization methods on both datasets and simulation, demonstrating that our method has better generalization ability. Our project is available at \url{https://sites.google.com/view/dg-traj-gen}.
\end{abstract}

\section{Introduction}
\begin{figure}[t]
\center 
\includegraphics[width=0.48\textwidth]{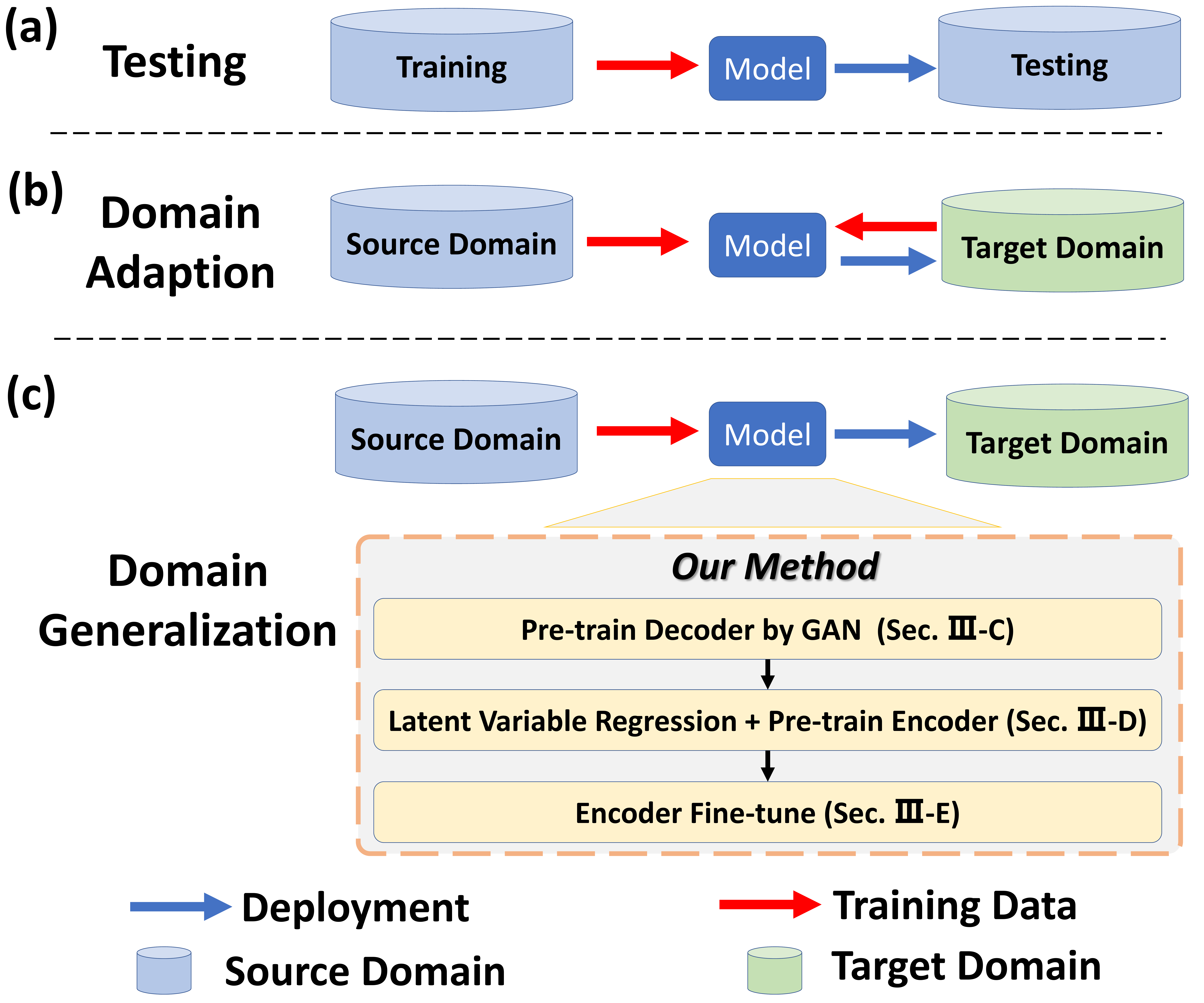}
\caption{Different frameworks for testing, domain adaption, and domain generalization method are shown in (a), (b), and (c) respectively. We propose a three-stage training approach to realize real domain generalized trajectory generation method for autonomous vehicles.}
\label{intro}
\end{figure}

Vision-based navigation is an appealing research topic in recent years. One of the approaches in vision-based navigation is learning-based trajectory generation from RGB images.
Most of the works are only validated on data from the same domain for training and testing, as shown in Fig.~\ref{intro}(a).
However, as a common challenge in learning-based algorithms, out-of-distribution (OOD) scenarios can lead these trajectory generation approaches to have poor generalization results and make dangerous decisions.

To allow the model to be trained on a specific dataset and transferred to a new scenario, a straightforward solution to deal with the OOD problem is \textit{domain adaptation (DA)} approach, which collects some data from the target domain to fine-tune the model, shown in Fig.~\ref{intro}(b). However, in the autonomous driving application scenario, a more realistic setting is that the target domain is unknown, so it is impossible to obtain the target domain data in advance. The problem dealing with such OOD scenario is introduced as \textit{domain generalization (DG)} \cite{zhou2021dgsurvey}, which aims to learn a model from source domains and generalize to any OOD target domain, shown in Fig.~\ref{intro}(c).

To realize domain generalization, one approach is to use labels of multi-domain training data \cite{li2018deep} \cite{ilse2020diva} to build an auxiliary task, trying to learn domain invariant data representations. However, domain labels are difficult to clearly define on most current driving datasets.
One approach to relax this problem is using ensemble learning \cite{filos2020can} to train a certain number of models with different training data and obtain a more robust planning result according to outputs of all models, while this approach needs a longer training time and inference time, making it resource-consuming. 
One appealing approach proposed recently is \textit{Invariant Risk Minimization (IRM)} \cite{arjovsky2019invariant}, which assumes the invariance of the feature-conditioned label distribution and aims to remove spurious correlations (i.e. dataset-specific biases) in a representation. However, most of the works using IRM are restricted to classification problems with simple datasets and linear classifiers.

In this paper, we proposed a trajectory generation method to real domain generalized visual navigation for autonomous vehicles by extending the IRM approach.
We construct an encoder-decoder network structure, where the decoder uses the method in \cite{wang2021imitation} to represent continuous trajectories.
In order to satisfy the constraint in the IRM problem, we use a Lagrangian form, which is the squared norm of the gradient of the trajectory generator.
It is difficult to solve this problem directly because the gradient norm term can hinder the optimization.
To solve this problem, we propose a three-stage training approach, which is shown in the lower part of Fig.~\ref{intro}(c): 1) We leverage an adversarial learning approach to train a trajectory generator as the decoder. 2) Based on the pre-trained decoder, we infer the latent variables corresponding to the trajectories, and pre-train the encoder by regressing the inferred latent variable. 3) We fix the decoder but fine-tune the encoder with the final trajectory loss.

We compare our proposed method with the state-of-the-art trajectory generation method and some recent domain generalization methods on both datasets and simulation, demonstrating that our method has better generalization ability. 
To the best of our knowledge, this is the first work to train driving trajectory generation models in one domain and directly transfer them to other domains.
To summarize, the main contributions of this paper include the following:
\begin{itemize}
    \item We formulate the domain generalization for driving trajectory generation problem as a non-linear IRM problem. And we propose a set of network training strategies for optimizing the non-linear IRM problem.
	\item We implement a trajectory generator with good domain generalization ability. We test our method on both datasets and simulation, showing that our method has a stronger generalization ability than others in both open-loop and closed-loop experiments.
\end{itemize}
\section{Related Works}
\subsection{Domain Generalization}
The goal of the domain generalization (DG) problem \cite{zhou2021dgsurvey} is to learn a model using data from the source domain and generalize to any out-of-distribution (OOD) target domain. Existing domain generalization approaches generally fall into the following groups:

\textbf{Domain-Adversarial Learning.} The goal of domain-adversarial learning \cite{li2018domain, li2018deep, carlucci2019hallucinating} is to align the distributions among different domains, and it formulates the distribution minimization problem through a minimax two-player game, without explicitly measuring the divergence between two probability distributions.

\textbf{Learning Disentangled Representations.} These approaches learn to represent the data as multiple features instead of a single domain-invariant feature and separate out the domain-specific parts \cite{ilse2020diva, xing2021domain}.

\textbf{Ensemble Learning.} It uses different splits of training data to learn multiple models with the same structure but different weights, which can boost the performance of a single model \cite{kahn2017uncertainty} \cite{tai2019visual}. Ensemble learning is effective to cope with OOD data with fewer constraints. However, these approaches require more computational resources, the training time and inference time of these approaches grow linearly with the number of models.

\textbf{Invariant Risk Minimization.} It was first proposed by Arjovsky et al. \cite{arjovsky2019invariant}, which aims to remove spurious correlations in a representation and ensure that the learned representation can lead to a minimal classification error over all source domains.
Recent new works inspired from IRM to address the OOD generalization problem include \cite{krueger2021out, jin2020domain, rosenfeld2020risks, ahuja2020invariant}. However, most of these approaches are validated only by doing classification tasks on toy datasets. We extend the IRM method by using non-linear models to do the trajectory generation task. And compared to other domain generalization methods, our proposed method does not rely on domain knowledge, pixel-reconstruction, or multiple models ensembling, making it easy to train and apply.

\subsection{OOD Generalization in Driving Policy Learning}
Recently, several works have investigated the problem of OOD generalization problem in driving policy learning. Zhang et al. \cite{zhang2021learning} proposed to use bisimulation metrics to learn robust latent representations which encode only the task-relevant information from observations in reinforcement learning. Unfortunately, there are still challenges in applying reinforcement learning methods to real-world driving tasks.
Filos et al. \cite{filos2020can} proposed an epistemic uncertainty-aware trajectory planning method by training an ensemble of density estimators and online optimizing the trajectory concerning the most pessimistic model. This approach achieves good results for OOD scenarios. However, model ensembling and online optimization will consume more computational resources and have a longer inference time, while in this paper, we only use one single model to implement OOD generalization.
\begin{figure*}[t]
\center
\includegraphics[width=0.95\textwidth]{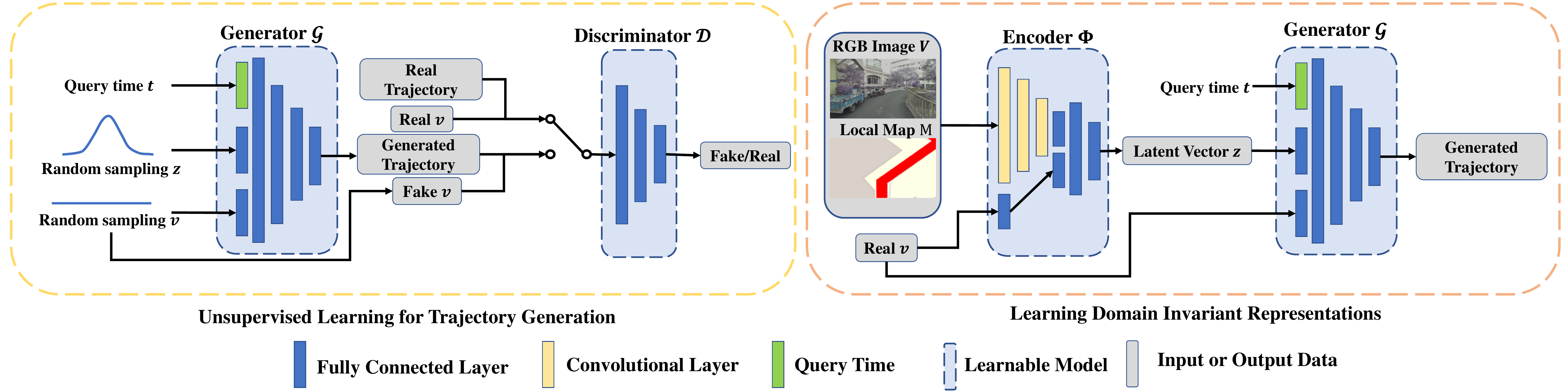}
\caption{
Network structure diagram of our proposed method. The generator $\mathcal{G}$ is trained with the discriminator $\mathcal{D}$ by an unsupervised adversarial learning approach as a decoder. Based on the pre-trained decoder, we infer the latent variables corresponding to the trajectories, and pre-train the encoder by regressing the inferred latent variable. Finally, we fix the decoder but fine-tune the encoder with the final trajectory loss.
The inputs of the encoder $\Phi$ include the RGB image $V$ which is acquired from the front-view camera, the local routing planning map $M$ which is cropped from an offline map according to the current low-cost GPS and IMU data, and the current speed $v$.
}
\label{fig:structure}
\end{figure*}
\section{Method}
\subsection{Background}
Most works in driving tasks use \textit{Empirical Risk Minimization (ERM)} principle, which assumes there is a joint probability distribution $P(x,y)$ over input data $X$ and label data $Y$, and the training data is drawn i.i.d. from $P(x,y)$. Given a loss function $L({\hat{y}},y)$ which measures how different the prediction $\hat{y}$ is from the ground truth $y$, ERM aims to find a hypothesis $h^*$ to minimize the empirical risk, which can be formulated as the following optimization problem:
\begin{equation}
h^* = \argmin_{h} \frac{1}{N} \sum_{i=1}^{N} L(h(x_i), y_i)
\end{equation}

However, when the joint probability distribution $P(x,y)$ varies in the testing data, ERM methods may yield poor generalization results. To overcome the distribution shift problem, as well as the absence of target domain data, domain generalization is introduced.

One of the important approaches in domain generalization is \textit{Invariant Risk Minimization} \cite{arjovsky2019invariant}, which assumes the invariance of the feature-conditioned label distribution $\mathbb{E}[y|\Phi(x)]$.
To find an approximate solution, a Lagrangian form is introduced \cite{rosenfeld2021the}:
\begin{equation}
\label{new-irm}
    \min_{\substack{\Phi, \mathcal{G}}}\sum_{e \in \mathcal{E}} \Big [\mathcal{R}^e(\Phi, \mathcal{G}) + \lambda \Vert \nabla_{\mathcal{G}} R^e(\Phi, \mathcal{G}) \Vert_2^2 \Big ]
\end{equation}
where $\Phi$ is a data representation, $\mathcal{G}$ is a classifier, $e$ is a kind of environment from the set $\mathcal{E}$, and $\lambda$ is a regularization parameter. The latter IRM penalty term constrains the classifier to be the optimal classifier and is not worse for each sample, avoiding the situation in ERM methods where some data errors may be very large. In the linear regression case, the analytical solution of the problem can be directly calculated, and the IRM problem can be simplified as \textit{IRMv1} \cite{arjovsky2019invariant}:
\begin{equation}
\label{irmv1}
    \min_{\substack{\Phi}}\sum_{e \in \mathcal{E}} \Big [\mathcal{R}^e(\Phi) + \lambda \Vert \nabla_{\mathcal{G} | \mathcal{G}=1.0} R^e(\Phi, \mathcal{G}) \Vert_2^2 \Big ]
\end{equation}
where $\Phi$ becomes the entire invariant predictor, $\mathcal{G}=1.0$ is a scalar and fixed ``dummy'' classifier.

\subsection{Problem Setup}
\label{problem-setup}
For the vision-based driving trajectory generation task, we leverage three sensor data as input of our method: the front-view RGB image $V$ which contains environmental information, the local route planning map $M$ which contains driving intention information as inputs, and the current speed $v$ of the vehicle. The local route planning map $M$ is cropped from an offline map based on the current pose from the low-cost GPS and inertial measurement unit (IMU), which is similar to \cite{amini2019variational} and \cite{ma2019towards}. Both images are resized to $400\times200$ and concatenated together as part of the encoder's input.

We assume that in the driving trajectory generation task, there is an encoder $\Phi$ which maps all the sensor data to the latent variable $\boldsymbol{z}$ and a trajectory generator (decoder) $\mathcal{G}$ which uses the latent variable $\boldsymbol{z}$, current speed $v$, and the query time $t$ to generate trajectory points. The expression for the encoder $\Phi$ can be written as:
\begin{equation}
\boldsymbol{z} = \Phi(V, M, v)
\end{equation}
and the expression for the trajectory generator $\mathcal{G}$ \cite{wang2021imitation} can be written as:
\begin{equation}
\label{traj-expr}
\boldsymbol{y}(t) = \mathcal{G}(t, \boldsymbol{z}, v)
\end{equation}
The network structure is shown in the right part of Fig.~\ref{fig:structure}.

The risk function used in this paper is:
\begin{equation}
\mathcal{R} = \sum \Big [ \Vert \boldsymbol{y}_\parallel - \hat{\boldsymbol{y}}_\parallel\Vert_2^2 + \alpha \Vert\boldsymbol{y}_{\perp} - \hat{\boldsymbol{y}}_{\perp}\Vert_2^2 \Big ]
\end{equation}
where $\boldsymbol{y}_\parallel$ and $\boldsymbol{y}_{\perp}$ are the generated longitudinal and lateral displacement respectively, $\hat{\boldsymbol{y}}_\parallel$ and $\hat{\boldsymbol{y}}_{\perp}$ are the ground truth longitudinal and lateral displacement respectively, $\alpha$ is a regularizer balancing between longitudinal loss and lateral loss, and we use $\alpha = 5$ in our experiments.

The most straightforward approach is to use the \textit{IRMv1} method to regress discrete trajectory points using a linear predictor, as mentioned in \cite{arjovsky2019invariant}. We call this method as \textit{Traj IRM}. In our subsequent experiments, we find that \textit{Traj IRM} method does not perform well enough in the trajectory generation task. Therefore, we introduce a non-linear decoder to improve the model representation ability. Refer to Eq.~\ref{new-irm}, we also consider using a Lagrangian form, and the non-linear IRM optimization problem becomes:
\begin{equation}
\label{non-linear-irm}
\min_{\theta, \omega} \sum_{e \in \mathcal{E}} \Big [ \mathcal{R}^e(\Phi_\theta, \mathcal{G}_w) + \lambda \Vert \nabla_{w}\, \mathcal{R}^e(\Phi_\theta, \mathcal{G}_w) \Vert_2^2 \Big ]
\end{equation}
where $\theta$ is the parameters of the encoder $\Phi$, and $w$ is the parameters of the decoder $\mathcal{G}$. The former term is the ERM term and the latter term is the IRM penalty term. It is difficult to solve this problem directly because the IRM penalty term can hinder the optimization.

Hence, we propose to use a three-stage approach to learn the IRM regularized trajectory generation: 1) We leverage an adversarial learning approach to train a trajectory generator as the decoder. 2) Based on the pre-trained decoder, we infer the latent variables corresponding to the trajectories and pre-train the encoder by regressing the inferred latent variable. 3) We fix the decoder but fine-tune the encoder with our proposed trajectory loss in an end-to-end manner.

\subsection{Unsupervised Learning for Trajectory Generation}
\textbf{Trajectory Representation.} In this paper, we use a non-linear decoder to generate continuous trajectories, which was proposed in \cite{wang2021imitation}.
In that work, the trajectory generator $\mathcal{G}$ takes three inputs: current speed $v$ as a condition, the latent variable $\boldsymbol{z}$ as a trajectory prior, and the query time $t$, and it outputs the trajectory point corresponding to the time $t$, which has the same expression as Eq.~\ref{traj-expr}.
High-order physical quantities such as velocity and acceleration can be obtained analytically by calculating the high-order partial derivatives of the outputs with respect to time $t$:
\begin{equation}
\begin{aligned}
v(t) = \frac{\partial \boldsymbol{y}(t)}{\partial t},
a(t) = \frac{\partial^2 \boldsymbol{y}(t)}{\partial t^2}
\label{devobs}
\end{aligned}
\end{equation}
In this paper, we follow this method to represent trajectories.

\textbf{Latent Action Space Learning.} 
\label{latent-learning}
Since the linear classifier of IRM can be solved analytically, the convergence of the classifier is not a concern, while the non-linear classifier has convergence problems. Therefore, it is important to train a good decoder without relying on high-dimensional inputs. We propose to use the unsupervised GAN to train the decoder. After training, the generator can be seen as a trained and converged decoder, which to some extent can be analogous to the classifier in linear IRM case, and we then can fix it in the process to optimize Eq.~\ref{non-linear-irm}.

Specifically, inspired by the multi-modal experiment in \textit{Conditional GAN} \cite{mirza2014conditional}, we sample the trajectory prior $\boldsymbol{z}$ from a standard Gaussian distribution $p_z(\boldsymbol{z})$, and a one-dimensional noise speed $\tilde{v}$ from a uniform distribution $p_{\tilde{v}}(\boldsymbol{\tilde{v}})$ between $0$ and the maximum speed as a condition:
\begin{equation}
\boldsymbol{z} \sim p_z(\boldsymbol{z}), \tilde{v} \sim p_{\tilde{v}}(\boldsymbol{\tilde{v}})
\end{equation}
And the trajectory generator $\mathcal{G}$ takes these variables to generate a fake trajectory:
\begin{equation}
\tilde{\boldsymbol{y}}(t) = \mathcal{G}(t, \boldsymbol{z}, \tilde{v})
\end{equation}
Then we use a time series with equal interval sampling and input to the trajectory function to obtain the corresponding discrete trajectory points. The discriminator network $\mathcal{D}$ takes these generated trajectory points $\tilde{\boldsymbol{y}}$ or ground truth trajectory points $\boldsymbol{y}$ as input, and determines whether it is sampled from the generator network or from the ground truth data.
The network structure is shown in the left part of Fig.~\ref{fig:structure}.
To improve the stability of the training process and prevent severe model collapse, we use WGAN-GP \cite{gulrajani2017improved} as our adversarial learning approach.
\subsection{Encoder Pre-training}
\label{subsec-encoder-pretrain}
We view the encoder pre-training task as a regression problem on the latent action space.
However, the problem is that there is no target latent variables $\hat{\boldsymbol{z}}$ to supervise the encoder training. Therefore, we infer the latent variables for each trajectory.
Referring to the Eq.~\ref{loss-fix}, we use the risk function along with the IRM loss. Since the latent variables are sampled from the standard Gaussian distribution in the training process of the GAN model, we also add a constraint on the norm of the latent variable $\boldsymbol{z}$ so that its distribution is as close as possible to the standard Gaussian distribution. The final loss function to obtain the target latent variable  $\hat{\boldsymbol{z}}$ is shown in Eq.~\ref{latent-reg}.
\begin{equation}
\label{latent-reg}
\begin{aligned}
\hat{\boldsymbol{z}} = \argmin_{\boldsymbol{z}} \sum_{e \in \mathcal{E}} \Big [ \mathcal{R}^e(\boldsymbol{z}, \mathcal{G}_{w_0}) &+ \lambda\Vert \nabla_{w|{w = w_0}}\, \mathcal{R}^e(\boldsymbol{z}, \mathcal{G}_w) \Vert^2 \\
&+ \lambda_2 \Vert \boldsymbol{z} \Vert^2_2 \Big ]
\end{aligned}
\end{equation}
where $w_0$ is the parameters pre-trained by GAN, $\lambda_2$ is a regularizer.
In practice, we use Adam optimizer with $0.1$ initial learning rate to optimize the latent variable $\hat{\boldsymbol{z}}$ as the target label data for supervised learning. Then we pre-train the encoder by regressing the latent variable $z$, and the loss function is:
\begin{equation}
\label{encoder-pretrain}
\theta_0 = \argmin_{\theta} \sum \Vert \Phi_\theta(V, M, v)-\hat{\boldsymbol{z}}\Vert^2_2
\end{equation}

After obtaining the latent variables $\hat{\boldsymbol{z}}$ from the multi-step optimization, a simple way is to follow the IRM approach to do linear regression on the latent space by using \textit{IRMv1} method \cite{arjovsky2019invariant}. However, we believe that it will introduce both the error of the latent variables used for supervision and the error of learning these latent variables, and our subsequent experiments show that this method works, but not very well. We call this method as \textit{Latent IRMv1}.

\subsection{End-to-End Training}
\label{e2e-train}
After pre-training the encoder, we train our model in an end-to-end manner.
In order to guarantee the feature-conditioned label distribution $\mathbb{E}[y|\Phi(x)]$ remains invariant, we propose to fix the parameters of the decoder $\mathcal{G}$ in the training process, and just fine-tune the encoder.
We use the GAN pre-trained parameters as the fixed parameters for the decoder, instead of random parameters, which can reduce the difficulty of optimization in the latent space and speed up training.

Therefore, the optimization problem in Eq.~\ref{non-linear-irm} is reduced to only search the parameters of the encoder:
\begin{equation}
\label{loss-fix}
\min_{\theta} \sum_{e \in \mathcal{E}} \Big [ \mathcal{R}^e(\Phi_\theta, \mathcal{G}_{w_0}) + \lambda \Vert \nabla_{w|{w = w_0}}\, \mathcal{R}^e(\Phi_\theta, \mathcal{G}_w) \Vert_2^2 \Big ]
\end{equation}
We call the loss function in Eq.~\ref{loss-fix} as \textit{Non-linear IRM (NIRM)} loss.

In our experiments, we also discuss the use of a decoder with random parameters, i.e., the decoder has not yet converged well, but we use it to train the model in the same way. We call this method \textit{Random NT+NIRM}. And our subsequent experiments demonstrate the poor performance of the model trained by this method.

\section{Experiments}
In this paper, we use different ablated models to validate the effectiveness of our proposed method on the open-source driving datasets and compare our method with the state-of-the-art trajectory generation method and recent domain generalization methods on the open-source driving datasets and the CARLA simulation.
\subsection{Dataset and Metrics}
We use three driving datasets to validate our method.

\textbf{Oxford Radar RobotCar} (RobotCar) \cite{barnes2020oxford} is a radar extension to the Oxford RobotCar Dataset\cite{maddern20171}, providing $280 km$ driving data around Oxford, UK. Since this dataset has simpler and practical data collection conditions, with limited geographic space and different collection times, we use this dataset as a training dataset.

\textbf{KITTI Raw Data} (KITTI) \cite{geiger2013vision} contains $6$ hours of traffic scenarios using a variety of sensor modalities. Since this dataset has more diverse driving scenarios, we use this dataset as a testing dataset.

\textbf{CARLA Dataset} is collected by a human driver for about $2$ hours in the CARLA \cite{dosovitskiy2017carla} simulation, with a speed limit of $30~km/h$ under different weather conditions. Since the cost of changing weather conditions in the simulation is very low and the driving trajectories are relatively simple, it is used as a testing dataset in the experiments.

\textbf{Metrics.} In this paper, we use average displacement error \cite{cai2020vtgnet}, which is the average Euclidean distance between the ground truth trajectory points and the generated trajectory points at the corresponding moment, to evaluate the performance of different methods.

\subsection{Ablation Study}
We validate the advantages of our proposed method by testing different ablated models from the original model as below:
\begin{itemize}
    \item \textit{E2E NT}: We directly train the model end-to-end in an ERM manner.
    \item \textit{E2E NT+NIRM}: We train the model end-to-end with \textit{NIRM} loss.
    \item \textit{Random NT+NIRM}: This method is proposed in Sec.~\ref{e2e-train}. We use a fixed neural trajectory generation model with random parameters instead of the parameters of the pre-trained GAN model.
    \item \textit{Traj IRM}: This method was proposed in \cite{arjovsky2019invariant}, and is mentioned in Sec.~\ref{problem-setup}. We use this method to regress the position coordinates of $16$ discrete trajectory points.
    \item \textit{Latent IRMv1}: This method is proposed in Sec.~\ref{subsec-encoder-pretrain}, which do linear regression on the latent space by using \textit{IRMv1} method.
    \item \textit{Ours}: The method proposed in this paper, which is trained by our proposed three-stage training approach.
\end{itemize}

\begin{table}[tb]
\center
\caption{Ablation study results of generalization from RobotCar Dataset to KITTI and CARLA Dataset. The three columns on the right are the average displacement error (m) on three different testing datasets. Lower metrics have better results.}
\label{ablation-table}
\begin{threeparttable}
\begin{tabular}{c@{}c@{\ \ }c@{\ \ }c@{\ \ \ }c@{\ \ \ }c@{\ \ \ }c@{\ \ \ }c}
\toprule
Algorithm & DFX & DPT & IRM & RobotCar* & KITTI & CARLA\\
\midrule
E2E NT\cite{wang2021imitation} &   &              &              & $\first{0.60}$ & $2.13$ & $1.36$\\
E2E NT+NIRM           &   &              & NIRM & $0.68$ & $2.06$ & $1.25$\\
Random NT+NIRM        &   $\checkmark$  &              & NIRM & $1.50$ & $2.32$ & $1.45$\\
Traj IRM    & $\checkmark$ &  & IRMv1 & $0.67$ &  $2.16$ & $1.31$\\
Latent IRMv1     &   $\checkmark$& $\checkmark$ & IRMv1 & $0.77$ &  $1.77$ & $1.02$\\
Ours                  &  $\checkmark$ & $\checkmark$ & NIRM & $0.85$ & $\first{1.70}$ & $\first{0.92}$\\
\bottomrule
\end{tabular}
\begin{tablenotes}
\item ``DFX'' means ``Decoder Fixed'', ``DPT'' means ``Decoder Pre-trained'', and ``IRM'' means ``With IRM''.
\end{tablenotes}
\end{threeparttable}
\end{table}

The ablation study results of transfer from RobotCar Dataset to KITTI Dataset and CARLA Dataset are shown in Tab.~\ref{ablation-table}.
Comparing with the results of the \textit{E2E NT} method, using IRM can improve model generalization ability, which illustrates the effectiveness of the IRM approach.
Comparing with the results of the \textit{E2E NT+NIRM} method and \textit{Random NT+NIRM} method, our method has a stronger generalization ability. Since the pre-trained decoder we use is a converged decoder with fixed parameters, we believe it is a correct analogy to the linear IRM case, so our method has a better performance.
Comparing the results of the \textit{Traj IRM} method, it shows that the decoder using a simple ``dummy'' predictor does not have enough model capacity for more complex problems such as trajectory generation for autonomous vehicles.
The second-best performing \textit{Latent IRMv1} is also analogous to the linear IRM case, but it only uses intermediate results as supervision, so there are fitting errors introduced, resulting in slightly worse performance.

\subsection{Comparative Study}
We compare our proposed method with the recent trajectory generation method which aims to overcome the OOD challenge and other domain generalization methods. To be fair, all methods use the same inputs as our proposed method. For the discrete trajectory generation methods, we use linear interpolation to get the trajectory points at the corresponding moment. For methods that require domain labels, we divide three datasets into $5$ domains respectively, according to the time of data collection. Note that our method does not require domain labels.

\textbf{End-to-End Neural Trajectory (E2E NT)} is a variant of \cite{wang2021imitation}, which has the same network structure and inputs as our proposed method. All its parameters are trained end-to-end with only $L_2$ loss. This method is treated as a baseline method in this paper.

\textbf{Robust Imitative Planning (RIP)} \cite{filos2020can} is one of the state-of-the-art methods for trajectory generation to overcome distribution shifts. We use the \textit{Worst Case Model (WCM)} over $5$ models and Adam \cite{kingma2014adam} optimizer for online optimization with $50$ steps and $0.1$ initial learning rate. According to the official code provided, the model output is $4$ discrete trajectory points, and the points at other moments are obtained using linear interpolation.

\textbf{MixStyle} \cite{zhou2021domain} is a method to make CNNs more domain-generalizable by mixing instance-level feature statistics of training samples across domains without using domain labels. We use this plugin in our end-to-end model without changing other settings for a fair comparison.

\textbf{Domain Invariant Variational Autoencoders (DIVA)} \cite{ilse2020diva} is a generative model that tackles the domain generalization problem by learning three independent latent subspaces, one for the domain, one for the class, and one for any residual variations. We extend this method to turn the classification task into a trajectory generation task, using the continuous trajectory generation model proposed in \cite{wang2021imitation}. This method needs domain labels in the training process.

\textbf{Domain-Adversarial Learning (DAL)} \cite{ganin2015unsupervised} methods leverage adversarial learning to allow the generator to extract domain invariant features. We use this method in our end-to-end model, and design a discriminator to determine whether two features come from the same domain. This method also needs domain labels in the training process.

\begin{figure}[t]
\center 
\includegraphics[width=0.45\textwidth]{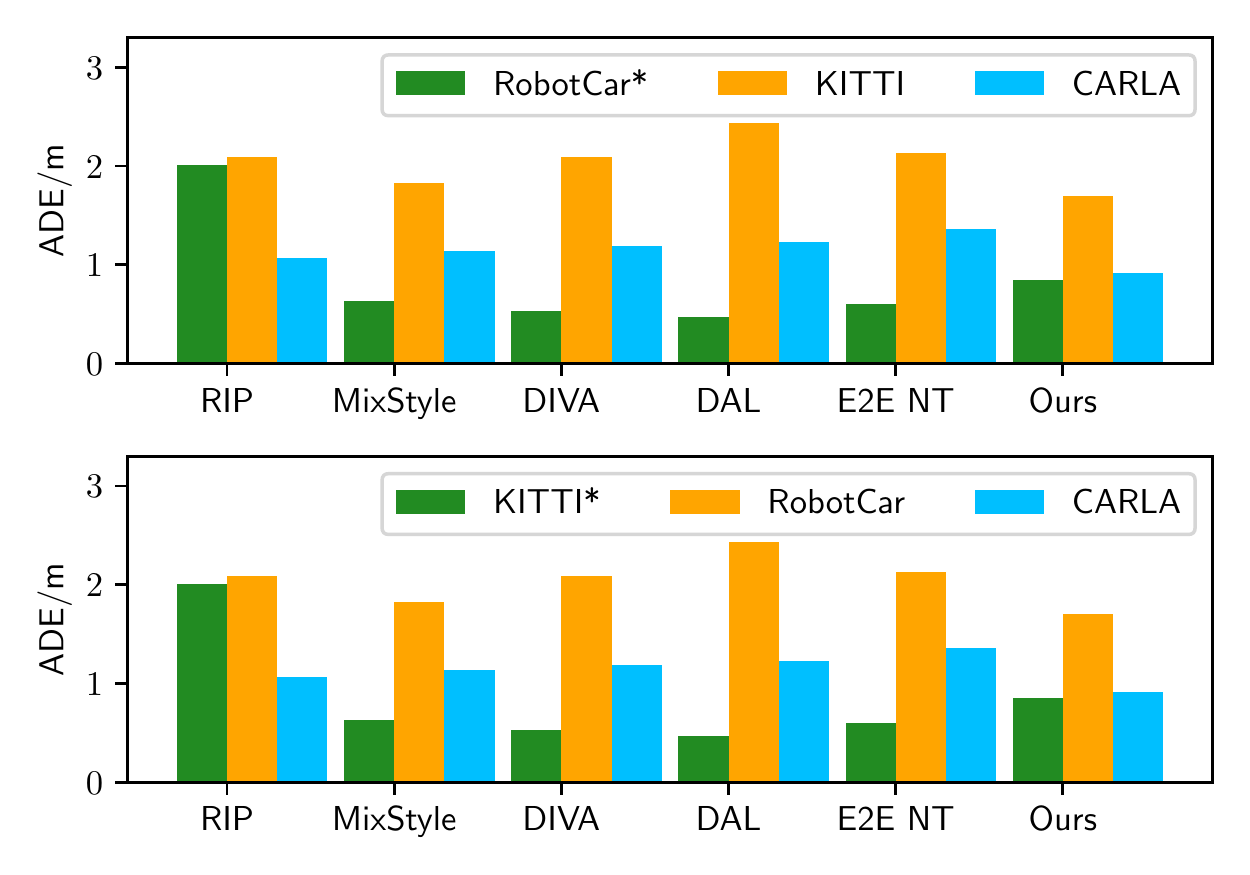}
\caption{Generalization performance (average displacement error in meters) on three different datasets. The models are trained on the dataset labeled by ``*'', and directly generalize to the testing dataset and other two target datasets.}
\label{comp}
\end{figure}

\subsection{Comparison Results on Datasets}
We implement our network by using PyTorch 1.6 with CUDA 10.2 and cuDNN 7.6.5 libraries. We use a batch size of $32$ and Adam \cite{kingma2014adam} optimizer with an initial learning rate of $0.0003$. All networks are trained on a PC with AMD 3700X CPU and NVIDIA RTX 2060 Super GPU.
We train all methods on training dataset until the models converge, and evaluate them on the testing datasets from both the same domain and other domians.\par
The generalization results of the different models are shown in Fig~\ref{comp}. In terms of generalization performance, our method outperforms other comparison methods on the testing datasets under different domains, which validates that our method has a stronger generalization ability.
The results show that \textit{MixStyle} and \textit{DIVA} these two domain generalization methods also have a great performance improvement compared to the \textit{E2E NT} baseline method. While the \textit{RIP} method and the \textit{DAL} method do not always show a stable generalization performance advantage. Under certain conditions, the performance of these two methods may be worse than that of the baseline method.

\begin{table}[tb]
\center
\caption{Closed-loop testing of success rate ($\%$, the first one of each item) and average speed ($m/s$, the second one of each item) in CARLA. Higher metrics have better results. Highest success rates are highlighted in bold font.}
\label{closed-loop}
\begin{tabular}{l@{\ \ \ \ }c@{\ \ \ \ }c@{\ \ \ \ }c@{\ \ \ \ }c}
\toprule
\multirow{2}{*}{Algorithm} & Clear & Wet Cloudy & Hard Rain & Heavy Fog\\
 & Noon & Sunset & Sunset & Morning\\
\midrule
RIP\cite{filos2020can} & $\third{86/3.6}$ & $\second{78/3.8}$ & $\second{82/3.7}$ & $\second{76/3.7}$\\
MixStyle\cite{zhou2021domain} & $\second{94/4.1}$ & $\third{77/4.9}$ & $\third{72/7.4}$ & $34/6.4$\\
DIVA\cite{ilse2020diva} & $79/9.6$ & $71/9.3$ & $54/8.7$ & $\third{39/7.5}$\\
DAL\cite{ganin2015unsupervised} & $79/4.0$ & $31/4.9$ & $27/4.7$ & $38/3.8$\\
E2E NT\cite{wang2021imitation} & $\first{100/8.6}$ & $50/4.5$ & $35/3.9$ & $31/9.9$\\
$\mathbf{Ours}$ & $\second{94/5.6}$ & $\first{82/6.8}$ & $\first{100/7.6}$ & $\first{79/5.7}$\\
\bottomrule
\end{tabular}
\end{table}

\subsection{Closed-loop Experiments in Simulation}
Since we want to train our driving model on the dataset collected in one environment and transfer this model to a new environment to implement driving tasks, evaluation on open-loop datasets is not enough to prove the effectiveness of our method. Therefore, we test our method with closed-loop visual navigation tasks in the CARLA \cite{dosovitskiy2017carla} 0.9.9.4 simulation and compare it with other methods.

\textbf{Experiments Setup.} We train models on RobotCar Dataset and transfer them in the CARLA simulator, testing the driving success rates under different driving tasks of different models.
We use the same vehicle and set random starting and target points in \textit{Town01} with four different weather conditions: \textit{Clear Noon}, \textit{Cloudy Sunset}, \textit{Hard Rain Sunset}, and \textit{Heavy Fog Morning} and two traffic condition: \textit{Empty} and \textit{with Dynamic Obstacles}, where the setting of obstacles is the same as its in the CARLA benchmark \cite{dosovitskiy2017carla}.

\begin{figure}[t]
\center 
\includegraphics[width=0.45\textwidth]{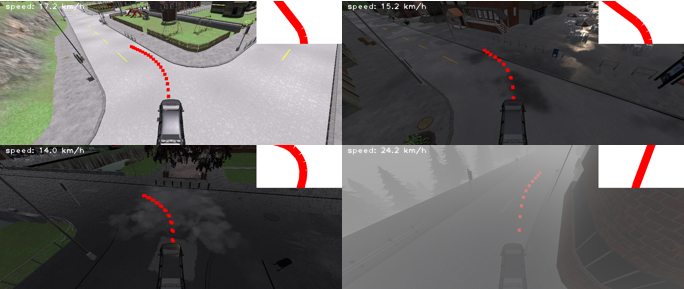}
\caption{Model generalization results of our method under four different weather conditions in CARLA. The model is only trained on RobotCar dataset and directly generalize to CARLA. The discrete red points are the generated trajectory points.}
\label{carla-gen}
\end{figure}

\textbf{Closed-loop Experiment Result.} The results are shown in Tab.~\ref{closed-loop} and Fig.~\ref{carla-gen}. Our method gets second place in success rate under \textit{Clear Noon} weather condition and has the highest success rate under all other weather conditions. Compared to the \textit{E2E NT} method, our method has high success rates in all weather conditions, while the success rates of \textit{E2E NT} method vary greatly under different weather conditions, which means that our method has a stronger model transfer ability to handle different weather conditions. Compared to the RIP method, our method has higher success rates and higher average speeds, while the \textit{RIP} method using \textit{Worst Case Model (WCM)} generates more conservative trajectories with low speed. Compared to \textit{MixStyle} and \textit{DIVA}, which have high success rates under the former three weather conditions, our method can also get a high success rate under \textit{Heavy Fog Morning} weather condition, where there is a close field of view and severe visual disturbance, which also validate our method has a stronger transfer ability.
\section{Conclusion}
In this paper, we propose a domain generalization method for vision-based driving trajectory generation for autonomous vehicles in urban environments, which can be seen as a solution to extend the IRM method in non-linear cases. We compare our proposed method with the state-of-the-art trajectory generation method and some recent domain generalization methods on both datasets and simulation, demonstrating that our method has better generalization ability.

\small{
\bibliography{reference}
}
\end{document}